\documentclass[pdflatex,sn-basic]{sn-jnl}

\usepackage[T1]{fontenc}
\usepackage[utf8]{inputenc}
\usepackage{amsmath,amssymb}
\usepackage{booktabs}
\usepackage{tabularx}
\usepackage{longtable}
\usepackage{multirow}
\usepackage{enumitem}
\usepackage{hyperref}
\usepackage{xcolor}
\usepackage{tikz}
\usetikzlibrary{arrows.meta,positioning,shapes.geometric,fit}

\title[From GPT-3 to GPT-5]{From GPT-3 to GPT-5: Mapping their capabilities, scope, limitations, and consequences}

\author*[1]{\fnm{Hina} \sur{Afridi}}
\author[2]{\fnm{Habib} \sur{Ullah}}
\author[3]{\fnm{Sultan Daud} \sur{Khan}}
\author[4, 5]{\fnm{Mohib} \sur{Ullah}}

\affil*[1]{\orgdiv{MediaFutures, Department of Information Science and Media Studies}, \orgname{University of Bergen}, \orgaddress{\city{Bergen}, \country{Norway}}}
\affil[2]{\orgdiv{Faculty of Science and Technology}, \orgname{Norwegian University of Life Sciences}, \orgaddress{\city{Ås}, \country{Norway}}}
\affil[3]{\orgdiv{Faculty of Computing and Information Technology}, \orgname{Sohar University}, \orgaddress{\city{Sohar}, \country{Oman}}}
\affil[4]{\orgdiv{Department of Computer Science (IDI)}, \orgname{Norwegian University of Science and Technology}, \orgaddress{\city{Gj\o vik}, \country{Norway}}}
\affil[5]{\orgdiv{School of Economics, Innovation and Technology}, \orgname{Kristiania University of Applied Sciences}, \orgaddress{\city{Oslo}, \country{Norway}}}

\abstract{%
We present the progress of the GPT family from GPT-3 through GPT-3.5, GPT-4, GPT-4 Turbo, GPT-4o, GPT-4.1, and the GPT-5 family. Our work is comparative rather than merely historical. We investigates how the family evolved in technical framing, user interaction, modality, deployment architecture, and governance viewpoint. The work focuses on five recurring themes: technical progression, capability changes, deployment shifts, persistent limitations, and downstream consequences. In term of research design, we consider official technical reports, system cards, API and model documentation, product announcements, release notes, and peer-reviewed secondary studies. A primary assertion is that later GPT generations should not be interpreted only as larger or more accurate language models. Instead, the family evolves from a scaled few-shot text predictor into a set of aligned, multimodal, tool-oriented, long-context, and increasingly workflow-integrated systems. This development complicates simple model-to-model comparison because product routing, tool access, safety tuning, and interface design become part of the effective system. Across generations, several limitations remain unchanged: hallucination, prompt sensitivity, benchmark fragility, uneven behavior across domains and populations, and incomplete public transparency about architecture and training. However, the family has evolved software development, educational practice, information work, interface design, and discussions of frontier-model governance. We infer that the transition from GPT-3 to GPT-5 is best understood not only as an improvement in model capability, but also as a broader reformulation of what a deployable AI system is, how it is evaluated, and where responsibility should be located when such systems are used at scale.
}

\keywords{GPT, large language models, foundation models, frontier models, multimodality, alignment, AI governance.}

\begin{document}

\maketitle

\section{Introduction}

The GPT family has emerged as one of the most publicly prominent trajectories through modern LLM history. GPT-3 demonstrated the public significance of large-scale few-shot learning; instruction-tuned and chat-focused followups reset user expectations of how to query language models; GPT-4 expanded the perceived upper bound of general-purpose model performance; GPT-4o brought real-time multimodal interactivity to the forefront; GPT-4.1 highlighted long context, coding, and tool use for developers; and the GPT-5 lineup publicized a more explicitly workflow-guided and routed system architecture \cite{Brown2020, Ouyang2022, OpenAI2023GPT4, OpenAI2024GPT4o, OpenAI2025GPT41, OpenAI2025GPT5SC}. The family therefore matters for at least three reasons.

First, it defined much of the contemporary deployment regime of foundation models. Beyond affecting benchmark culture, the GPT family helped entrench API-mediated access, model-tiering, alignment layers, safety-reporting via system cards, and commercialization tactics that involve productizing model capabilities as reusable services \cite{Bommasani2022, OpenAI2024GPT4oSC, OpenAI2026Models}. Second, it altered conceptions of the nature of user-to-LLM interaction. Lauded as a prompt-completion engine for in-context learning, subsequent models shifted the focus toward dialogic assistance, multimodal interaction, and tool-augmented interfaces. Third, these engineering developments are starting to impact myriad sectors besides model development  \cite{Brown2020, Ouyang2022, OpenAI2024GPT4o}. The GPT series has affected research and development priorities, software engineering workflows, educational testing methodologies, interface design principles, knowledge work and content creation professions, media industries, and debates around safety, transparency, and oversight \cite{Floridi2020, Shahriar2024, Fachada2025}.


It is challenging to provide a review of the GPT family since the public documentation is non-homogeneous. GPT-3 and GPT-4 are presented in considerable research-style artifacts, whereas GPT-3.5 Turbo, GPT-4 Turbo, GPT-4.1, and GPT-5.x are documented through a mixture of papers, model pages, release notes, and system cards rather than fully comparable technical reports \cite{Brown2020, OpenAI2023GPT4, OpenAI2026GPT35Turbo, OpenAI2026GPT4Turbo, OpenAI2025GPT41, OpenAI2025GPT5SC}. Moreover, later generations are increasingly mixed up with routing logic, tools, and product interfaces, which means that ``the model'' and ``the deployed system'' are no longer easy to separate. We address these challenges by treating the GPT family as an evolutionary sequence of public AI systems rather than as a list of disconnected reviews. We aim to review successive generations along their technical profile, capability envelope, interaction style, disclosed limits, and consequences in use. We do not attempt to reconstruct undisclosed architectures or training methods from unverified sources. When official documentation leaves gaps, we treat those gaps as part of how modern AI models are governed in practice. We present how the GPT family has evolved technically and functionally from GPT‑3 to GPT‑5, examining what changed across generations in modality, context handling, tool use, and interaction design. We also highlight which limitations remained despite major improvements in alignment, instruction following, and benchmark performance. We also present the economic, social, and institutional consequences that emerged as these models became more conversational, multimodal, and widely deployable.

The paper is orgnized in the following sections. Section \ref{scoppe} mentions about the scope, sources, and methodology, Section \ref{termme} describes terminology and taxonomy, Section \ref{hissto} presents  historical evolution, Section \ref{techh} presents technical profile of the GPT family, Section \ref{cappat} highlights
capability comparison, and Section \ref{intter} presents interaction paradigm shift. The paper further describes limits and failure modes in Section \ref{limmit}, 
alignment, safety, and governance in Section \ref{alignnnt}, economic and infrastructural consequences in Section \ref{eccon}, societal and professional consequences in Section \ref{soccia}, changes across generations in Section \ref{synnt}, open challenges and future research directions in Section \ref{oppen},
and conclusion in Section \ref{concnn}.

\begin{figure*}[t]
    \centering
    \includegraphics[width=13cm,height=6cm]{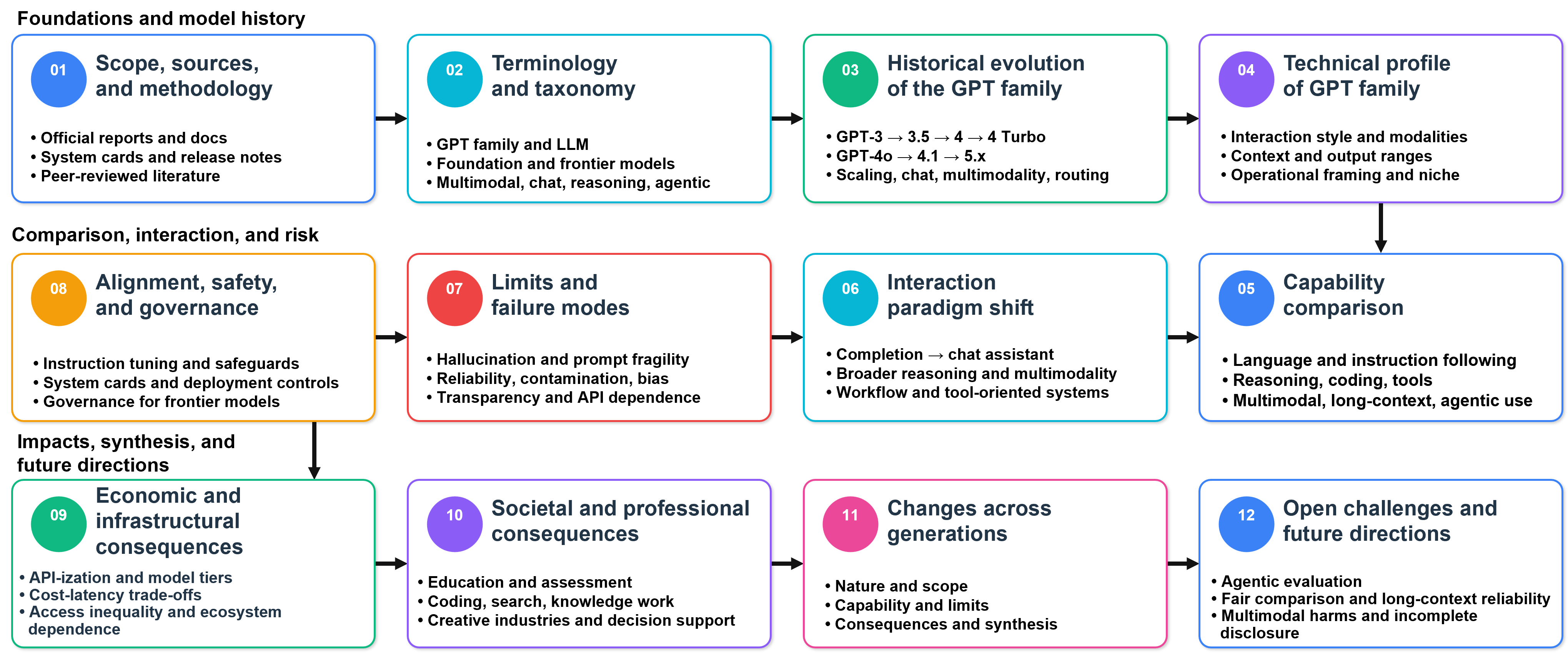}
    \caption{The overall structure of our work is divided into foundations and model history, comparison, interaction, and risk, impacts and future directions. }
    \label{fllowdiagram}
\end{figure*}

A graphical representation of the overall structure of our work is presented in Fig. \ref{fllowdiagram}, highlighting the aforementioned sections.

\section{Scope, sources, and methodology}
\label{scoppe}

We compiled information that is publicly available from the following five categories of sources: (i) technical reports, (ii) system cards, (iii) API/model documentation, (iv) product announcements / release notes, and (v) secondary literature. The primary sources are particularly valuable since later generations of GPTs have been documented inconsistently across artifacts. While GPT-3 and GPT-4 models are anchored by research publications \cite{Brown2020, OpenAI2023GPT4}, GPT-4o and GPT-5 have system cards that emphasize system safety and deployment considerations \cite{OpenAI2024GPT4oSC, OpenAI2025GPT5SC}. Meanwhile, GPT-3.5 Turbo, GPT-4 Turbo, GPT-4.1, and GPT-5.4 have model pages that discuss modalities, context length, max tokens limits, and expected use cases \cite{OpenAI2026GPT35Turbo, OpenAI2026GPT4Turbo, OpenAI2026GPT41Model, OpenAI2026GPT54}. Peer-reviewed research papers fall into two categories of usage. Some are used directly to analyze specific generations of GPTs, or comparisons to humans or peers in narrow use cases: ophthalmology, futuristic projection, multimodal analysis, and experiment automation  \cite{Shahriar2024, Fachada2025}. Others are used to broadly frame concepts such as foundation models, alignment, or social connotations of systems like GPTs \cite{Bommasani2022, Floridi2020}. Reference material provided by users of GPT-4 Turbo, GPT-4o, GPT-4.1, GPT-5, and generalized comparisons between GPT versions also inform this paper, but are not leaned on heavily as singular sources of information. They are viewed through the lens of triangulation with other methods. Lastly, this review is not designed to quantify into a single scorecard determining ``best model.'' Rather, it is meant to consistently compare how the public face of each model changed over time. The dimensions of analysis include: (i) style of interaction, (ii) modality, (iii) context and output capacity, (iv) tool use and chain of thought capabilities, (v) reported strengths, (vi) reported weaknesses, and (vii) impacts on society.

\section{Terminology and taxonomy}
\label{termme}

A consistent vocabulary is necessary because the GPT family extents research artifacts, API products, and integrated assistant systems.

\paragraph{GPT model family} The GPT model family points to the sequence of publicly documented OpenAI systems and models from GPT-3 onward that are outlined as generative pre-trained transformer \cite{X1} descendants, including chat-optimized, multimodal, and routed family variants \cite{Brown2020, OpenAI2025GPT5SC}.

\paragraph{Large language model (LLM)} An LLM is a large-scale model \cite{X2} trained on language-centric data to predict or generate text, often with transfer to diverse downstream problems. In the GPT context, this wide term remains useful even after the family extends beyond text-only interaction because text generation is still a key part of how the AI interacts with users \cite{Brown2020, Shahriar2024}.

\paragraph{Foundation model} Following widely used terminology, a foundation model \cite{X3} is a general-purpose model trained on large data at scale and adapted to many tasks and domains \cite{Bommasani2022}. GPT models fit this description.

\paragraph{Frontier model} A frontier model is not a completely different type of AI, it is a very powerful AI model that sits at the top of what is currently available to the public. OpenAI itself calls its latest GPT-5.4 model a frontier model, designed for demanding professional tasks \cite{OpenAI2026GPT54, OpenAI2026Models}.

\paragraph{Multimodal model} A multimodal model \cite{X4} accepts, reasons over, or produces more than one modality, such as text, images, audio, or video. GPT-4 introduced multimodal input in research documentation, and GPT-4o was publicly presented as a model that can fully handle text, images, and audio all at once \cite{OpenAI2023GPT4, OpenAI2024GPT4o, OpenAI2024GPT4oSC}.

\paragraph{Chat model} Unlike earlier AI models that simply finished a sentence, chat models are designed to have real conversations and follow instructions. GPT-3.5 Turbo is an example of this, and it became very popular when released to the public \cite{OpenAI2026GPT35Turbo, Ouyang2022}.

\paragraph{Reasoning model} In OpenAI's current products, reasoning models \cite{X5} are AI models that can control how much thinking effort they put into answering a question. For example, GPT-5.4 lets users adjust the level of reasoning effort through its API \cite{OpenAI2026GPT54}. In this paper, the term 'reasoning model' is used to describe this feature, not to make any claims about whether AI truly thinks.

\paragraph{Agentic workflow / tool-using system} It points to an AI system that can call tools, manage long contexts, interact with external resources, and execute multi-step workflows with minimal guidance. GPT-4.1 is presented as more effective for powering agents, and GPT-5.4 is positioned for agentic, coding, and professional workflows \cite{OpenAI2025GPT41, OpenAI2026GPT54}.

\begin{figure*}[t]
    \centering
    \includegraphics[width=13cm,height=4.5cm]{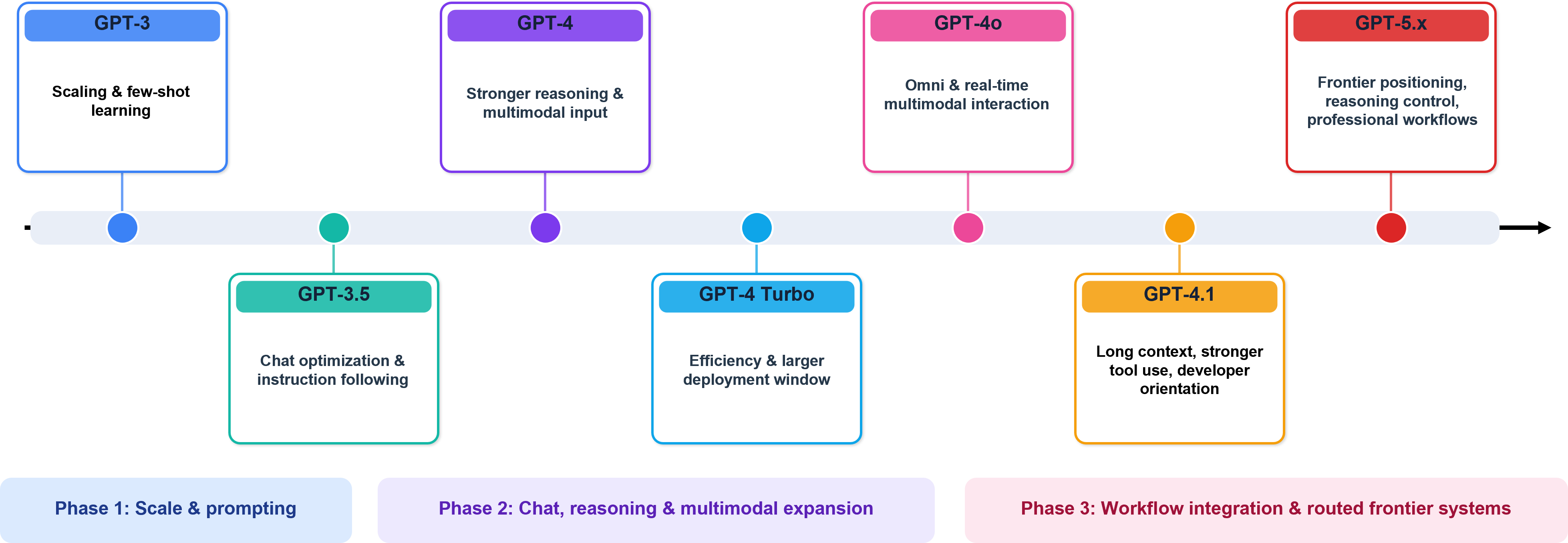}
    \caption{We show the evolution of the GPT family from GPT-3 to GPT-5.x. The figure emphasizes shifts in scale and prompting, chat and reasoning, workflow integration and frontier systems.}
    \label{gpptfamily}
\end{figure*} 

\section{Historical evolution of the GPT family}
\label{hissto}

In this section, we describe GPT-3, GPT-3.5, GPT-4, GPT-4 Turbo, GPT-4o, GPT-4.1, and GPT-5.x. We depict this historical evolution in Fig. \ref{gpptfamily}.

\subsection{GPT-3: Scaling and few-shot learning}

GPT-3 demonstrated a significant concept that scale itself could drive capabilities broadly across task-agnostic few-shot \cite{X6} performance. Brown et al. \cite{Brown2020} presented GPT-3 as a 175B-parameter autoregressive language model, and demonstrated that wide-scale pre-training followed by in-context prompting could achieve competent performance at translation, question answering, and other language tasks. They used a context window of 2,048 tokens. Owing to these public discussions, GPT-3 became representation of a style of research: one general model to many tasks, few gradient updates at deployment time. But GPT-3 was not, at its core, meant to be deployed as a chat system. The user was still responsible for supplying prompts, demonstrations, and formatting patterns to encourage the system toward desired behaviors. Capability and usability were separate considerations. GPT-3 made in-context learning seem practical, but it also made prompt design an essential engineering practice. Historically, GPT-3 should be seen less as a finished assistant and more as a scalable text engine whose few-shot behavior would reshape the research landscape of later generations.

\subsection{GPT-3.5: Chat optimization and instruction following}

One might instead consider GPT-3.5 as the point where public-facing generative AI models went from being completion engines to chat assistants centered around language models like GPT. Alignment research resulting in InstructGPT demonstrated that models fine-tuned with human feedback were better able to align with user intent, fact generations, and reduce untruthful text generation compared to raw GPT-3-style prompts \cite{Ouyang2022}. GPT-3.5 Turbo brought this effort to effect in a public-facing API model focused on chat use cases while still being able to perform as a strong code and non-chat model \cite{OpenAI2026GPT35Turbo}. The change from GPT-3 that one finds most salient, however, is not necessarily this incremental improvement. It is the shift in user interface logic. Interaction with the model changes from ``show the model some pattern'' to ``tell the assistant what you want.'' This adjustment was socially significant in that it opened up accessibility of these models beyond those who deeply understood prompt design. However, GPT-3.5 is less transparent to researchers than its predecessor. Documentation is spread across alignment studies, product announcement posts, and model documentation rather than a flagship technical document.

\subsection{GPT-4: Stronger reasoning and multimodal input}

GPT-4 was a new public benchmark in performance and reporting. OpenAI announced GPT-4 as a large multimodal model taking image \cite{xx5} and text as input and creating text output \cite{OpenAI2023GPT4}. The report highlighted predictable scaling, widespread benchmark improvements, and better instruction-following after its post-training phase. GPT-4 notably surpassed GPT-3.5 on many professional and academic tests, including a simulated bar exam score in the upper 10\% of test participants \cite{OpenAI2023GPT4}. Importantly, GPT-4 set a precedent for public style of reporting capabilities. Benchmarking comparisons, safety disclosures, and outright statements of limitations were presented together. The model's hallucinations, short context window, and inability to learn from experience after deployment were all stated. That combination of transparency about strengths and weaknesses would come to influence later releases of GPT, as well as the industry at large.

\subsection{GPT-4 Turbo: Efficiency and a larger practical deployment window}

One can consider GPT-4 Turbo as the first major public effort to wring out broad GPT-4-level capability into a shape more practically deployable at economic API scale. Contemporary OpenAI model documentation described GPT-4 Turbo as essentially a cheaper and better GPT-4, capable of accepting text and image input and text output with a 128K context window \cite{OpenAI2026GPT4Turbo}. On a practical level, this mattered because it meant large-document workflows, retrieval-heavy prompting, and other long-session use cases suddenly became much more practical. What GPT-4 Turbo did not do conceptually was redefine the basic paradigm of interaction. Instead, GPT-4 Turbo expanded the deployment envelope. This increased the reach of GPT-4 era capabilities by reducing cost and raising context limits. The tradeoff here is that GPT-4 Turbo will likely be remembered historically less for any particular conceptual leap and more as an important optimization along the trajectory of the model family.

\subsection{GPT-4o: Omni and real-time multimodal interaction}

GPT-4o \cite{X7} represented an even clearer transition in concept. OpenAI initially described it as an omni model that can take text, audio, image \cite{Xx4}, and video input in any combination and generate text, audio, and image outputs, including responses to audio inputs with latencies nearly as fast as realtime human conversation \cite{OpenAI2024GPT4o}. The GPT-4o system card elaborated on this by describing it as an end-to-end autoregressive omni modal model trained on text, vision, and audio data which also required new attention to emerging voice-specific risks \cite{OpenAI2024GPT4oSC}. The key shift is in how the assistant operates. GPT-4o is positioned not as just a more capable text model with some modalities appended, but rather as a fundamentally more natural system for human-computer interaction. The speech channel becomes central. So is multimodal understanding and real-time conversation. Both the utility and risk profiles expand. The system card raises concerns about copyrighted material, speaker identification, illicit use of voice cloning, persuasion, and other risks unique to the added modalities not highlighted in previous iterations of GPT. Secondary evaluation studies likewise emphasize that GPT-4o extends comparison criteria beyond text-only reasoning to language, vision, speech, and multimodal performance, while still exhibiting variability and modality-specific weaknesses \cite{Shahriar2024}.

\subsection{GPT-4.1: Long context, stronger tool use, and developer orientation}

GPT-4.1 is significantly importance since its public reputation revolves around developers. GPT-4.1, GPT-4.1 mini, and GPT-4.1 nano were announced by OpenAI primarily as API models with coding, instruction following, and long context improvements, with the ability to process context up to one million tokens \cite{OpenAI2025GPT41}. According to the available GPT-4.1 model page, it is the smartest model being released that does not include a reasoning step. It has a context window of 1,047,576 tokens, a maximum output token limit of 32,768, wide ranging instruction-following and tool-calling capability, and low latency \cite{OpenAI2026GPT41Model}. The release of GPT-4.1 was a significant shift from multimodal assistant to workflow component. Where many earlier iterations of ChatGPT have been marketed towards consumers as a chatbot destination, GPT-4.1 is uniquely being marketed towards developers as backend infrastructure for building code editors, working in repositories, analyzing long documents, and agent-like systems more broadly \cite{OpenAI2025GPT41}. Peer reviewed benchmarks also suggest this focus. \cite{Fachada2025} presented particularly strong GPT-4.1 performance in Python code generation given unfamiliar libraries, while also highlighting remaining needs for verification and careful prompt design \cite{Fachada2025}.

\subsection{GPT-5.x: Frontier positioning, reasoning control, and professional workflows}

The GPT-5 family represents different models. Public technical documentation frames GPT-5 as a system, comprised of a fast model, a deeper reasoning model, and a router \cite{X8} that dispatches to these models based on conversation type, expected complexity, tool requirements, and user intent explicitly stated within the conversation \cite{OpenAI2025GPT5SC}. According to OpenAI's official documentation, GPT-5.4 is a top-tier model developed for complex professional tasks. It allows users to adjust how much reasoning effort it applies, can handle around one million tokens of text, and can generate up to 128,000 tokens in a single response. It also supports a range of tools, including functions, internet search, file search, and computer use \cite{OpenAI2026GPT54, OpenAI2026Models}. This is important because the way GPT-5 is publicly presented deliberately makes the definition of model unclear. Beyond framing GPT-5 in terms of a base capability profile, OpenAI documents the model's routing, tiers of functionality \cite{X9}, safety training regime via safe-completions, and specialized professional workflows for writing, coding, and healthcare \cite{OpenAI2025GPT5SC}. The GPT-5 family marks a point at which deployment architecture, safety mechanisms, and reasoning options are subsumed into what is publicly considered part of the model itself.

\section{Technical profile of GPT family}
\label{techh}

Table~\ref{tab:master_comparison} presents a compact cross-generation layout of GPT models. Several patterns stand out as can be seen in the Table. First, there is a growth in the scope of interaction: from completion engine, to chatbot, multimodal, and workflow agent. Next is the expansion of accessible context, which has increased from 2K tokens to approximately 1M tokens in modern long-context successors \cite{Brown2020, OpenAI2026GPT41Model, OpenAI2026GPT54}. Third is the evolving role of tools. The use of tools did not seem to be an important part of how the public understood or used GPT-3 and GPT-4. However, for GPT-4.1 and GPT-5.4, tools appear to be a key part of both what these models can do and what they are designed to be used for  \cite{OpenAI2026GPT41Model}\cite{OpenAI2026GPT54}. Finally, the table shows an interesting pattern: newer models are used in more real-world applications, but at the same time, less detailed information about them is shared publicly for research purposes.

\begin{table}[ht]
\centering
\caption{We present the differences in the GPT family in terms of interaction style, modality framing, context window, maximum output, and operational framing.}
\scriptsize
\setlength{\tabcolsep}{5pt}
\begin{tabularx}{\textwidth}{p{1.1cm} p{1.5cm} p{1.6cm} p{1.7cm} p{1.6cm} X}
\toprule
\textbf{Model} & \textbf{Interaction style} & \textbf{Modality framing} & \textbf{Context window} & \textbf{Max output} & \textbf{Operational framing} \\
\midrule
GPT-3 & Prompt-completion, few-shot in context & Text in, text out & 2,048 tokens & Within context budget & Few-shot prompting; no chat-first framing; publicly presented as a large autoregressive language model. \\
GPT-3.5 Turbo & Chat-optimized dialog & Text in, text out & 16,385 tokens & 4,096 tokens & Chat Completions optimization; useful for code and non-chat tasks; publicly framed as a legacy chat model for instruction-following. \\
GPT-4 Turbo & Assistant model with long context & Text and image in, text out & 128,000 tokens & 4,096 tokens & More deployable GPT-4 variant for long-context workflows; publicly framed as a cheaper, better GPT-4 and now legacy relative to newer models. \\
GPT-4o & Real-time omni assistant & text, audio, image, video input; text, audio, image output & 128,000 tokens & 16,384 tokens & Multimodal interaction with later specialized realtime/audio variants and broader safety focus. \\
GPT-4.1 & Developer-oriented non-reasoning model & Text and image in, text out & 1,047,576 tokens & 32,768 tokens & Explicit tool calling, long context, and agent-enabling workflows; publicly framed as non-reasoning model for coding, instruction following, and large-context tasks. \\
GPT-5.x / GPT-5.4 & Routed frontier workflow system / flagship API model & Text and image in, text out in current flagship API & 1,050,000 tokens (GPT-5.4) & 128,000 tokens (GPT-5.4) & Configurable reasoning effort, tools, routed fast-vs-thinking behavior, and safe-completions; publicly framed as a frontier model for agentic, coding, and professional workflows. \\
\bottomrule
\end{tabularx}
\label{tab:master_comparison}
\end{table}

\section{Capability comparison}
\label{cappat}

\begin{figure}[t]
    \centering
    \includegraphics[width=13cm,height=4cm]{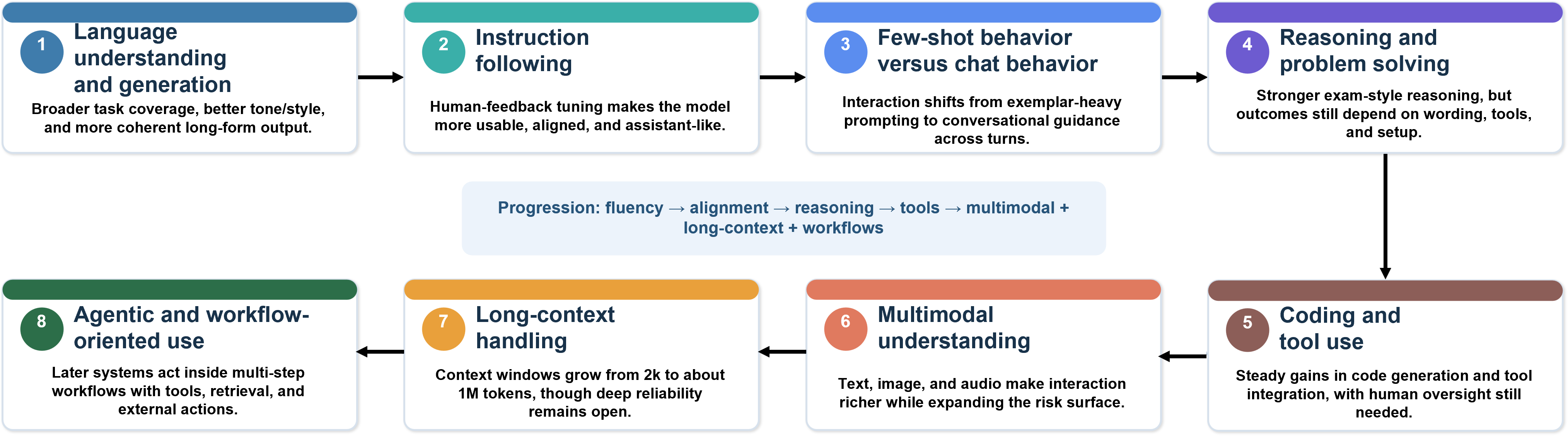}
    \caption{A flow view of how the GPT family expands from fluent text generation to multimodal, long-context, tool-using workflows considering their capabilities.}
    \label{capppcomparison}
\end{figure}

The GPT family can be compared not only by technical profile but by the kind of work each model enables or performs. In Fig. \ref{capppcomparison}, we present how the family expands from fluent text generation to multimodal, long-context, and tool-using workflows. 

\subsection{Language understanding and generation}

GPT-3 demonstrated that given a text prompt, an adequately scaled model could be applied to a variety of tasks \cite{Brown2020}\cite{X10}. GPT-4 takes dramatic strides in accuracy on benchmark tests like massive multitask language understanding (MMLU) \cite{X11} and HumanEval \cite{OpenAI2023GPT4}\cite{X12}. It should be noted that these measures of language capability are not simply text generation, measured by something like readability fluency. Over successive releases in the GPT series, capability has improved not just at producing correct answers but appropriate tone and style for the user \cite{X13}, generalizing across subjects, and presenting longer responses in coherent formats. Outside of benchmarking, some researchers have explained why this matters. \cite{Floridi2020} noted that GPT-3 was significant not because it was smart but because it could produce convincing copious amounts of humanlike text that could be persuasive. Successor models build massively on this capability, better sentences, and closer and closer to exactly what the user desires.

\subsection{Instruction following}

Instruction following \cite{X13} is likely the single biggest improvement between models in the GPT family. InstructGPT demonstrated that fine-tuning a model with human feedback could result in a model that followed user instructions better than vastly larger models while also being more truthful and less likely to generate untruthful content \cite{Ouyang2022}. GPT-3.5 and the instruction tuned chat models \cite{X15} implemented this concept. GPT-4 improved upon this further, and GPT-4.1 had instruction following explicitly stated as a primary objective. OpenAI benchmarked GPT-4.1 as outperforming GPT-4o on instruction following \cite{OpenAI2025GPT41}. This is important because instruction following is considered a basic capability. It is necessary for the model to serve as a reliable assistant, utilize tools properly, perform multi-step tasks, and safely align its behavior. While a model may be impressive at generating fluent text but not follow instructions well, it is difficult to use safely and responsibly.

\subsection{Few-shot behavior versus chat behavior}

One feature that was often noticed about GPT-3 was that it could learn from a few examples provided directly in the prompt \cite{Brown2020}. This technique is known as few-shot learning \cite{X16}. GPT-3.5 and successors retained this capability but they prioritized conversation instead. What this means in practice is that one no longer have to carefully provide examples to get good performance, one can just ask a question, reformulate it, provide more information, or change topics over the course of many turns. There's a subtle trade-off here. Making a system more conversational tends to make it easier to use but makes it harder to tell why one is getting a good answer, is it because of the model's inherent abilities, its training on human feedback, retrieved information from external knowledge, or the context of the conversation? It is even harder to determine this in systems like GPT-5 where the ultimate answer one gets may depend on hidden choices the system makes internally, such as which model to use or how much effort to reason with \cite{OpenAI2025GPT5SC}.

\subsection{Reasoning and problem solving}

GPT-4 represented the largest advancement in general reasoning within the GPT series thus far, as measured by standardized testing and exam performance \cite{OpenAI2023GPT4}. Both GPT-4.1 and GPT-5 further advanced reasoning capabilities but in divergent directions. GPT-4.1 emphasized code generation, long-context comprehension, and more grounded tool usage \cite{OpenAI2025GPT41}. GPT-5.4 improved upon this further by allowing users to control the effort the model applies for reasoning tasks, transitioning this improvement from an invisible background feature to a user-controllable setting \cite{OpenAI2026GPT54}. Reasoning ability is still far from perfect or guaranteed. Scores on testing can vary significantly based on wording, available tools, and even evaluation methodology. OpenAI has noted that some of their results have varied based on prompt or tool usage \cite{OpenAI2025GPT41}. All claims about these models should be taken as rough estimates, not exact statements of fact.

\subsection{Coding and tool use}

Coding \cite{X17} is an area where improvements to GPTs over time have been especially straightforward. GPT-4 performed much better on coding tasks than GPT-3.5, according to OpenAI's technical write-up \cite{OpenAI2023GPT4}. GPT-4.1 was then presented as a strong coding model, succeeding on popular coding benchmarks \cite{OpenAI2025GPT41}. Finally, highlighting of coding ability is one of the things that GPT-5 papers use to showcase the family's real-world strengths \cite{OpenAI2025GPT5SC}. Research outside of OpenAI reaches similar conclusions, but with important notes of caution. One paper found GPT-4.1 to be surprisingly competent at writing full code using libraries it had not been trained on. However, the researchers also cautioned that careful prompting and rigorous testing were still required \cite{Fachada2025}. This is consistent with the trend across the whole GPT family: the models have steadily become more useful at coding, but human oversight is still required.

\subsection{Multimodal understanding}

GPT-4 added image input capabilities, but it was not until GPT-4o that working with text, images, and audio became something of a model’s stronger ability \cite{OpenAI2023GPT4, OpenAI2024GPT4o}. OpenAI’s system card describes GPT-4o as having particular strengths with image and audio inputs, while its press announcement stressed how the model can respond to users in real time using all three input modalities \cite{OpenAI2024GPT4o, OpenAI2024GPT4oSC}. Researchers have separately evaluated GPT-4o’s performance with text, vision \cite{X18}\cite{X19}\cite{X20}, speech, and multimodal inputs, observing impressive capabilities while flagging inconsistent performance on unclear audio and ambiguous images \cite{Shahriar2024}. The key point is that multimodal input support fundamentally alters the risk profile of a model. Once one allows voice and image inputs as primary modes of interaction with the model, one can no longer consider things like speaker identification, voice synthesis without consent, and misattribute image-based inferences as edge cases \cite{OpenAI2024GPT4oSC}.

\subsection{Long-context handling}

The ability to handle long texts has become increasingly important across the GPT family. GPT-3 could only process a small amount of text at once, which limited what it could do in a single interaction \cite{Brown2020}. GPT-4 Turbo expanded this significantly, allowing up to 128k tokens of text \cite{OpenAI2026GPT4Turbo}. GPT-4.1 pushed this even further, supporting around one million tokens and claiming better understanding of large amounts of text \cite{OpenAI2025GPT41, OpenAI2026GPT41Model}. GPT-5.4 continues to offer this million-token capacity \cite{OpenAI2026GPT54}.
However, being able to handle long texts does not automatically mean the model reasons well across all of them. OpenAI highlights improved support and comprehension, but does not guarantee reliable performance in every situation involving long documents or multiple sources. How well these models truly reason over very long contexts remains an open question that researchers are still working to properly evaluate.

\subsection{Agentic and workflow-oriented use}

In its later stages, the GPT family has become increasingly focused on completing real-world tasks and workflows. GPT-4.1 was explicitly connected to building AI agents and using external tools \cite{OpenAI2025GPT41}. GPT-5.4 is described as a powerful model designed for complex workflows involving coding, professional tasks, and autonomous action, with support for tools such as web search, file search, and computer use \cite{OpenAI2026GPT54, OpenAI2026Models}. This marks a clear shift, from a model that simply answers questions to one that actively participates in and carries out multi-step tasks. This shift matters because it changes where things can go wrong. Errors may now come not just from the model itself, but also from how it uses tools, the quality of information it retrieves, how different components are coordinated, or how the system is presented to the user. As a result, later GPT models are much harder to evaluate using traditional text-based tests alone.

\section{Interaction paradigm shift}
\label{intter}

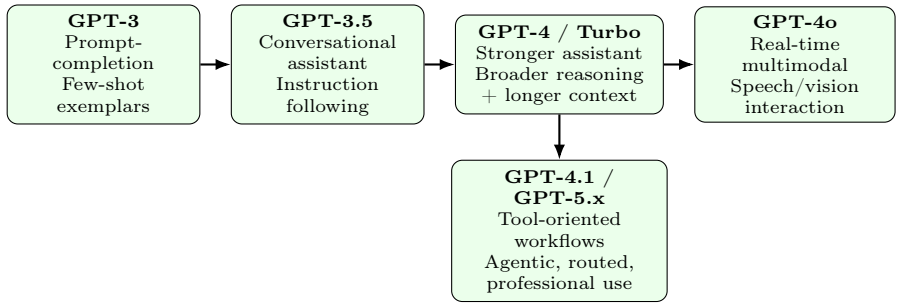
\begin{figure}[ht]
\centering
\begin{tikzpicture}[node distance=6mm and 4mm, every node/.style={font=\footnotesize}]
    \node[draw, rounded corners, fill=green!8, align=center, text width=2.3cm] (a) {\textbf{GPT-3}\\Prompt-completion\\Few-shot exemplars};
    \node[draw, rounded corners, fill=green!8, align=center, text width=2.3cm, right=of a] (b) {\textbf{GPT-3.5}\\Conversational assistant\\Instruction following};
    \node[draw, rounded corners, fill=green!8, align=center, text width=2.5cm, right=of b] (c) {\textbf{GPT-4 / Turbo}\\Stronger assistant\\Broader reasoning + longer context};
    \node[draw, rounded corners, fill=green!8, align=center, text width=2.4cm, right=of c] (d) {\textbf{GPT-4o}\\Real-time multimodal\\Speech/vision interaction};
    \node[draw, rounded corners, fill=green!8, align=center, text width=2.6cm, below=of c] (e) {\textbf{GPT-4.1 / GPT-5.x}\\Tool-oriented workflows\\Agentic, routed, professional use};

    \draw[-{Latex[length=2mm]}, thick] (a) -- (b);
    \draw[-{Latex[length=2mm]}, thick] (b) -- (c);
    \draw[-{Latex[length=2mm]}, thick] (c) -- (d);
    \draw[-{Latex[length=2mm]}, thick] (c) -- (e);
\end{tikzpicture}
\caption{Interaction paradigm shift across the GPT family. The later stages are not merely bigger chatbots; they increasingly function as multimodal and workflow-oriented systems.}
\label{fig:paradigm}
\end{figure}

The most important change in the GPT family is not only quantitative improvement but a change in interaction paradigm. We depicted this transtion in Fig.~\ref{fig:paradigm} representing prompt completion, conversational assistant, broader reasoning, multimodal interaction, and agentic workflows.

Each generation of the GPT family was built with a different kind of user interaction. GPT-3 assumed the user would carefully prepare a text prompt. GPT-3.5 assumed back-and-forth conversation. GPT-4 and GPT-4 Turbo considered the model would help with a broader range of complex tasks. GPT-4o assumed that users might interact through speech and images \cite{x21}\cite{x22} in real time. GPT-4.1 and GPT-5 assume the model may get involve in longer tasks, use external tools, and handle complex multi-step situations \cite{Brown2020, OpenAI2024GPT4o, OpenAI2025GPT41, OpenAI2025GPT5SC}. This shift has three major implications. First, it changes the users' expectations, we have moved from expecting blocks of text, to expecting assistance completing tasks. Second, it increases the complexity of evaluation, we care about more than just whether the response is correct, we care about the flow of the interaction, whether the model remains focused on task, and how it recovers from errors. Third, it mixes up responsibility for a given response. In a system that routes requests, allows for tool usage, and runs safety filters, the response seen by the user is produced by a stack of technologies working in conjunction, not simply one model making one decision.

\section{Limits and failure modes}
\label{limmit}

Despite substantial progress, several limitations recur across generations including hallucination, prompt sensitivity, reliability, benchmark contamination, bias and uneven performance, transparency, and depedence vendor-controlled APIs. We will briefly discuss each of them.

\subsection{Hallucination and factual instability}

Hallucination refers to the model generating a statement that is incorrect, invented, or unjustified. For example, hallucinations can include made-up references, incorrect dates or details, events that never happened, and high-confidence answers that should otherwise be withheld or deferred with a statement that the model does not know. Factual instability is the umbrella term. It encompasses any tendency for a models factual output to vary across prompts, runs, or contexts. A factually unstable model may get a question right when asked one way and partially (or fully) wrong when asked another way. It may give two competing answers when the context is subtly changed. Hallucination is one example of how these models can fail on facts; instability is the broader accuracy problem, also encompassing sensitivity to wording and poorly-calibrated confidence. This is how a model can have lower headline hallucination rates and still fail to be reliable in practice on some kinds of factual queries. That interpretation fits both the GPT-4 documentation \cite{OpenAI2023GPT4}, which acknowledges hallucination-related limitations, and the GPT-5 materials \cite{OpenAI2025GPT5SC}, which report lower hallucination rates without claiming the problem is solved. Therefore, hallucinations remain a fundamental challenge even as newer models improve \cite{x23}.

\subsection{Prompt sensitivity and specification fragility}

Prompt sensitivity \cite{x24} refers to substantial changes in a model's behavior when making equivalent tasks with changes to the wording, ordering, or formatting of the prompt. Specification fragility is the larger issue that a model may break when provided with a task specification that is underspecified, implicit, slightly ambiguous, or operationalized through an involved interaction structure. Put another way, the human user may know what they want, but the system may behave very differently due to what gets made into the specification through prompt phrasing, turn-taking structure, output format/schema, tool permissions, or agent configuration details. Prompt sensitivity was made widely visible through GPT-3 because few-shot performance varied drastically based on prompt phrasing, example ordering, and included textual cues. However, later instruction-tuned systems \cite{x25} do not eliminate this behavior. It simply manifests in other ways. Later models may not require carefully handcrafted exemplars but may behave sensitively to changes in phrasing, turn-taking structure, requested output format, implicit evaluation rubric, or available tools \cite{Brown2020, Ouyang2022, OpenAI2025GPT41}. This is especially crucial to consider when making claims about benchmark performance that depend on agent setup.

\subsection{Reliability under long reasoning chains}

Long reasoning chain \cite{x26, x27} reliability refers to the model's ability to maintain correctness, consistency, and alignment with the task at hand throughout many intermediate reasoning steps. This is particularly important when solving a problem with multiple stages, in a long context window, or where the correct answer requires maintaining multiple constraints simultaneously. It's not sufficient for each individual step of the model's reasoning to sound coherent or correct. The important question is whether the entirety of the chain is valid. A language model can produce fluent reasoning for many steps and still ultimately fail due to dropping an earlier constraint, misapplying a rule in the middle of its reasoning, or allowing a small local error to compound into an incorrect final answer. Longer reasoning traces or larger context windows do not necessarily mean stable solution quality. GPT-4 itself noted that it "may still make reasoning errors and is not fully reliable" for long reasoning chain tasks. GPT-5 also states that newer models use even smarter reasoning techniques by being trained to "think before they answer". Both of these points imply there has been progress in reasoning techniques but that it does not ensure correctness for every long horizon task.

\subsection{Benchmark contamination and comparability concerns}

Benchmark contamination \cite{x28, x29} refers to cases where some benchmark tasks, or highly similar variants of them, were included in the model’s training data or pre-training mixture. Benchmark contamination can artificially inflate benchmark scores since a model may partly rely on memorization instead of generalization. Comparability concerns refer to the broader issue that scores between models or generations may not be strictly comparable even when contamination is avoided or corrected for. This is because changing models often comes with changes to evaluation conditions: benchmarks used may differ, prompt styles may differ, chain-of-thought prompting \cite{Wei2022} may be used for some tasks but not others, tool access may or may not be added, internal evaluation pipelines may change, and so on. In fact, the GPT-4 report highlights these issues by using different benchmarks with different shot settings and by using chain-of-thought prompting for GSM-8K, showing that model score is partly dependent on evaluation setup. The GPT-5 system card page makes a similar point from another angle, noting that comparison values from live models may differ from launch-time values and drawing a distinction between legacy standard evaluations and newer, more difficult production benchmarks with lower scores that should not be interpreted as simple regression.

While a reported score increase can of course indicate real improvements in modeling, training, or alignment capabilities, it can also be influenced by changes to benchmark fit that don’t involve memorization (e.g., cleaner benchmarks, better prompting, stronger tool scaffolding, updated graders), or by using an entirely different benchmark culture. For these reasons, a later model may appear better not just because it is intrinsically better, but also because the ecosystem around it has changed.

\subsection{Bias and uneven performance}

Bias \cite{x30, x31} refers to situations where the model's outputs systematically replicate or reinforce socially objectionable patterns, stereotypes, exclusions, or skewed representations. Bias can manifest in many ways, whether by associating certain groups with undesirable attributes, generating culturally specific defaults, centering some identities while marking others as other or exceptional, or producing content that is incorrect, toxic, or unfairly differential across demographic, linguistic, or cultural groups. Uneven performance \cite{x32} refers to variability in the quality of a model's outputs across users, tasks, groups, languages, modalities, or other factors of interest. A model may have strong performance on average, yet perform noticeably worse for certain accents or languages, visual environments or domain-specific topics, or uncertain inputs. Put differently, uneven performance is about unequal distribution of capability while bias is about systematic skew that carries social or ethical weight. Bias and uneven quality are areas of significant concern. GPT-4 and GPT-4o safety documents address bias in terms of disparities, representational harms, and other uneven social impacts \cite{OpenAI2024GPT4oSC}. Multimodal systems can inherit and amplify such impacts via voice, image \cite{x33, x34}, and language.

\subsection{Transparency limits and incomplete disclosure}

Transparency limits \cite{x35, x36} refers to the boundaries placed on what the model developer is willing to make public about the workings of the system. This includes model architecture, model size, training data composition, data filtering and refinement practices, training compute, fine-tuning procedures, evaluation protocols, and deployment safeguards. The technical report of GPT-4 does a lot to disclose useful information about the capabilities, limitations, and safety research and engineering around the model. However, it does not go into the details about model architecture, parameter count, what hardware was used, training compute, how the dataset was constructed, and how the model was trained. Incomplete disclosure refers to the more specific problem that the published documentation does not include details that researchers, auditors, or third-party evaluators would typically require to fully recreate, verify, benchmark, or audit the model in a rigorous manner. Transparency limits are the broad status of being partially open about how the model works, while incomplete disclosure are the specific details omitted while under that status. The GPT-4 technical report mentions that information is being withheld for competitive and safety concerns. This is transparent in one sense, but gives the public only a partial picture of how the model was developed and trained.

Public transparency declines in some respects as product sophistication increases. GPT-3 provided a large research paper; GPT-4 provided a major technical report while withholding architecture and training-compute details; GPT-4o, GPT-4.1, and GPT-5.x are documented through a more fragmented mixture of system cards, launch posts, and model pages \cite{Brown2020, OpenAI2023GPT4, OpenAI2024GPT4oSC, OpenAI2025GPT41, OpenAI2025GPT5SC}. This matters because incomplete disclosure has methodological consequences. The lack of publicly available information about how these models are changing makes it difficult for researchers to know if improvements are coming from better modeling techniques or if training budgets are changing, if datasets are being better curated, if more effort is being put into post-training details, if alignment is improving due to the model or the prompt change, if tool scaffolding is providing the gains, or if live system updates are being made. Incomplete disclosure hurts reproducibility, auditability, and gives us less confidence in cross-generation comparisons.

\subsection{Dependence on vendor-controlled APIs}

Dependence on APIs controlled by vendors means that access to models passes through provider-hosted layers of software rather than being completely self-hosted with respect to model weights and compute infrastructure. Users and developers do not experience the model in isolation, then. They experience the model in conjunction with the vendor’s decisions about their platform: What model alias is mapped to which snapshot? Does a router send traffic to one submodel versus another? Which tools are activated? What rate limits are in effect? Have a model or endpoint been deprecated or completely shut down? OpenAI describes their models as being served via the Responses API and SDKs, further describes the GPT-5 system as routing in real-time based on conversation type, complexity, tool needs, and inferred user intent, lists aliases, snapshots, and tool availability on individual model pages. Therefore, the behavior of any model can depend on aliases, routing decisions, deprecations, safety filters, and tool policies outside the user's direct control \cite{OpenAI2026Models, OpenAI2026GPT54}. The resulting dependence is infrastructural as well as technical.

We provided the summary of the limits and failure modes across GPT generations in terms of hallucination, prompt sensitivity, reliability, benchmark contamination, bias, transparence and dependence on APIs in Table \ref{tab:limits_failure_modes_compact}.

\begin{table}[htbp]
\centering
\caption{Limits and failure modes across GPT generations in terms of hallucination, prompt sensitivity, reliability, benchmark contamination, bias, transparence and dependence on APIs.}
\scriptsize
\setlength{\tabcolsep}{4pt}
\begin{tabularx}{\linewidth}{p{3.2cm} p{5.2cm} X}
\toprule
\textbf{Failure mode} & \textbf{Brief meaning} & \textbf{GPT-family implication} \\
\midrule

Hallucination and factual instability 
& Hallucination is the generation of false, invented, or unsupported claims. Factual instability is the broader tendency of factual answers to vary across prompts, runs, or contexts. 
& Later GPT models reduce some factual errors, but the problem is not eliminated. GPT-4 acknowledges hallucination-related limitations, while GPT-5 reports improvement without claiming full resolution. \\

Prompt sensitivity and specification fragility 
& Model behavior can change substantially with small differences in wording, ordering, formatting, turn structure, or tool configuration. 
& GPT-3 made prompt sensitivity highly visible in few-shot prompting, and later instruction-tuned models still remain sensitive to how tasks are specified and scaffolded. \\

Reliability under long reasoning chains 
& The model may produce fluent multi-step reasoning but still fail by dropping constraints or compounding small intermediate errors. 
& Longer reasoning traces and larger context windows do not guarantee correctness.  \\

Benchmark contamination and comparability concerns 
& Benchmark contamination occurs when evaluation items, or close variants, appear in training data. Comparability concerns arise when cross-model comparisons use different prompts, tools, benchmarks, or evaluation pipelines. 
& Reported gains may reflect real modeling progress, but may also partly reflect benchmark fit, prompting style, tool scaffolding, or changing evaluation conditions. \\

Bias and uneven performance 
& Bias refers to systematic socially harmful skew, stereotypes, or exclusion. Uneven performance means model quality is not uniform across groups, languages, tasks, or modalities. 
& GPT-4 and GPT-4o safety materials discuss disparities and representational harms, and multimodal systems can extend these concerns through voice, image, and multilingual interaction. \\

Transparency limits and incomplete disclosure 
& Transparency limits are boundaries on what developers reveal publicly. Incomplete disclosure means key technical and evaluative details needed for replication or auditing are missing. 
& Later GPT generations are increasingly documented through partial technical reports, system cards, launch posts, and model pages, limiting transparency. \\

Dependence on vendor-controlled APIs 
& Access is mediated through provider-controlled APIs rather than fully self-hosted model weights and infrastructure. 
& Model behavior can depend on aliases, routing, tool availability, rate limits, deprecations, and safety policies outside the user's direct control. \\

\bottomrule
\end{tabularx}
\label{tab:limits_failure_modes_compact}
\end{table}

\section{Alignment, safety, and governance}
\label{alignnnt}

Alignment means configuring the model’s behavior to follow intended human goals, instructions, and norms \cite{x37}. Concretely, this usually means getting the model to be more helpful, follow legitimate instructions more strictly, avoid undesirable behaviors, and adhere to policy constraints. Safety means mitigating potential harms from use and misuse of the model \cite{x38}. Concretely, this covers preventing unsafe outputs, resisting jailbreak and prompt injection, limiting unsafe advice given by the model, monitoring and modeling unsafe capabilities, and building system-level safety guards. Governance \cite{x39, x40} is the more organizationally and macroscopic level: how the models are managed through rules, oversight and enforcement processes, access controls, deployment decisions, and accountability mechanisms.

Alignment has transitioned from being an afterthought to being a major and defining feature of how GPT models are developed. InstructGPT presented that human feedback could materially improve usefulness and reduce untruthful output relative to GPT-3 baselines \cite{Ouyang2022}. GPT-4 utilized wider safety reporting and domain expert red teaming \cite{OpenAI2023GPT4}. GPT-4o stretched the safety line to speech, vision \cite{X41, X42}, and multimodal risks, including persuasion and voice-related harms \cite{OpenAI2024GPT4oSC}. GPT-5 then highlighted safe-completions, instruction hierarchy, jailbreak resistance, and preparedness categories in its public safety framing \cite{OpenAI2025GPT5SC}. Several governance implications follow.

To begin with, the family increasingly recognizes that model safety cannot be lessened to pre-training data filtering alone. GPT-4o system documentation noted that many effective protections take place after pre-training through post-training, product safeguards, and enforcement layers \cite{OpenAI2024GPT4oSC}. Second, system cards have become an important, if imperfect, governance aspect. They do not replace for independent auditing, but they create a semi-standardized genre for reporting risks and mitigations.

subsequently, alignment remains a trade-off rather than a solved objective. A system can become more useful, more compliant with harmless instructions, and less weak in some safety categories while still failing in others or introducing new failure modes such as sycophancy, over-deference, or ambiguous boundary-setting \cite{OpenAI2025GPT5SC}. Governance discussions should therefore focus not only on model refusal or non-refusal, but on the quality and social adequacy of safe responses.

Finally, frontier governance becomes increasingly relevant as the family moves toward routed professional systems and agentic workflows \cite{X51, X52}. When models are positioned for coding \cite{X53}, health \cite{X54}, research assistance \cite{X55}, or other consequential domains, governance questions shift from ``can the model answer this?'' to ``under what deployment conditions should it be trusted, audited, rate-limited, or supervised?''

\section{Economic and infrastructural consequences}
\label{eccon}

Economic and infrastructural consequences refer to the macro effects of deploying GPT-family models on markets, accessibility, product form factors, and technical reliance as well as accuracy \cite{X56}. Economic consequences are effects on pricing tiers, commercial models, labor processes, competitive differentiation, and unequal access that result from advanced language-model capability being commercialized as a managed service. OpenAI’s API documentation describes the model family as a tiered product with separate price, latency, and capability tiers. That is itself an economic pattern: model quality is no longer just a demonstration or a research development; it’s also a metered utility that companies purchase, optimize for, and build products around. Infrastructural consequences are patterns of technical dependency introduced into the delivery and management stack for AI. Rather than downloading a single static flagship model and running it on our own hardware, users interact with a managed provider platform offering routing, model snapshots, rate limiting, toolchains, added safety layers, and deprecation schedules.

\begin{figure}[t]
    \centering
    \includegraphics[width=13cm,height=6cm]{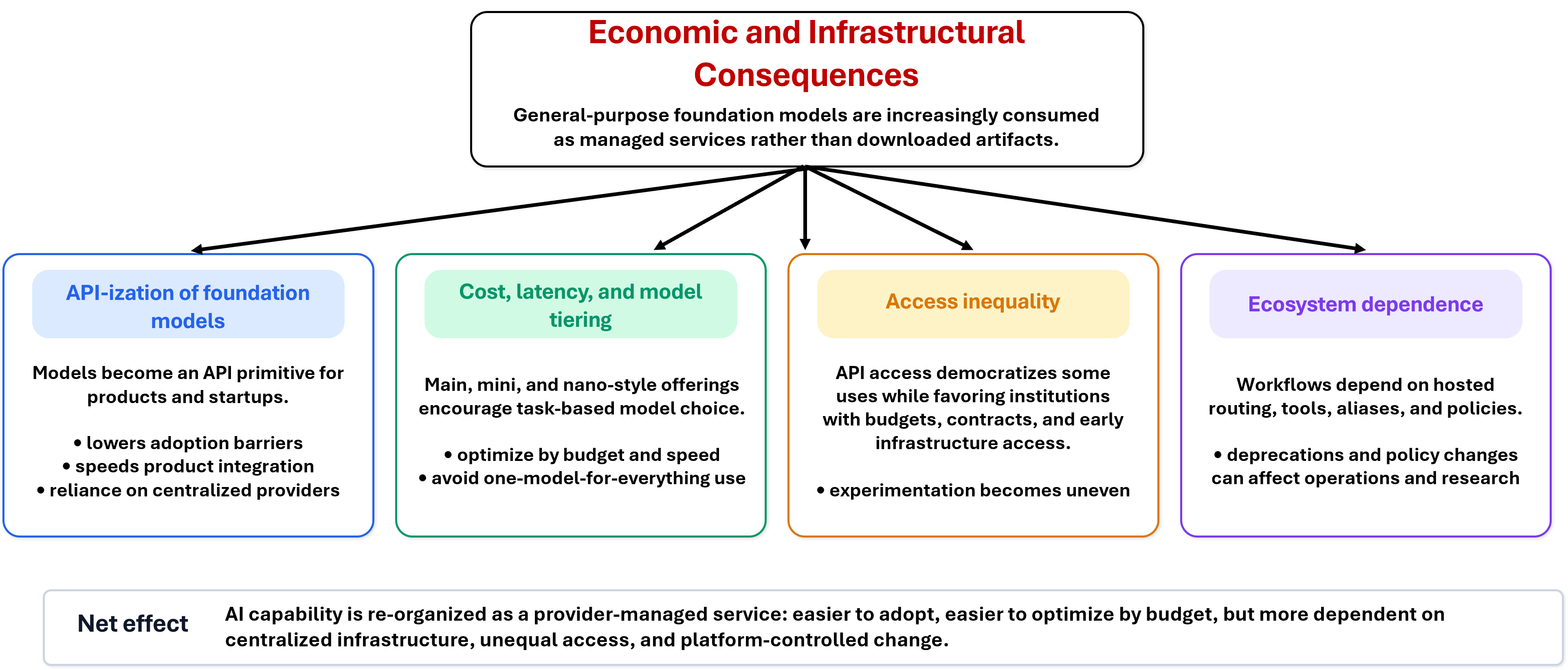}
    \caption{Economic and infrastructural consequences: GPTs capability is re-organized as a provider-managed service: easier to adopt, easier to optimize by budget, but more dependent on centralized infrastructure, unequal access, and platform-controlled change.}
    \label{ecconomicconse}
\end{figure}

The extensive growth of the GPT family helped standardize a commercial model in which cutting-edge capability is accessed via APIs, managed routing services, and tiered lines of models rather than distributed via open release of leading weights \cite{OpenAI2026Models}. This carries practical tradeoffs of incurring usage fees, locking in to a vendor’s model tiers, and ceding continued control of performance upgrades, endpoint behavior, and access changes to a service provider. It also lowers the entry barrier for startups and enterprises to develop products that use cutting-edge models without needing resources to train from scratch, which is empowering from an economic standpoint while also increasing lock-in. In fact, this has several consequences presented in Fig. \ref{ecconomicconse} and described below.

\subsection{API-ization of foundation models}

Models from the GPT family brought widespread attention to consuming a general-purpose foundation model as an API primitive instead of a concrete artifact to download. This reduces barriers to adoption for startups and product teams but heightens reliance on centralized providers \cite{Bommasani2022, OpenAI2026Models}.

\subsection{Cost, latency, and model tiering}

Scaling up the GPT family has also meant moving away from presenting it as a one-size fits all model. OpenAI now presents different flavors of models that tradeoff intelligence, speed, and price differently, via main, mini, and nano models or some analogous categorization \cite{OpenAI2025GPT41, OpenAI2026Models}. This is not just a commercial decision. It also impacts how developers use them by incentivizing users to select the most appropriate model for the task at hand, instead of using one model to do everything.

\subsection{Access inequality}

API-mediated access democratizes certain uses while exacerbating others. Researchers, institutions, and those with developer budgets or enterprise contracts (or proximity to platform insiders) have access to experiment on the latest models and tools longer and at scale than most independent researchers or cash-strapped institutions. In this way, access starts to look less like downloading parameters and more like access to being 'inside' the infrastructure.

\subsection{Ecosystem dependence}

As later iterations of GPT systems lean further into blackboxed routing, safety tuning, and hosted tools, external developers may end up building critical pieces of their workflows on top of interfaces that they do not maintain. Deprecations, alias changes, or policy updates can thus have operational and research impact \cite{X58}. This poses distinct but significant consequences for startups, enterprises, and academia alike \cite{X59}.

\begin{figure}[t]
    \centering
    \includegraphics[width=11cm,height=8cm]{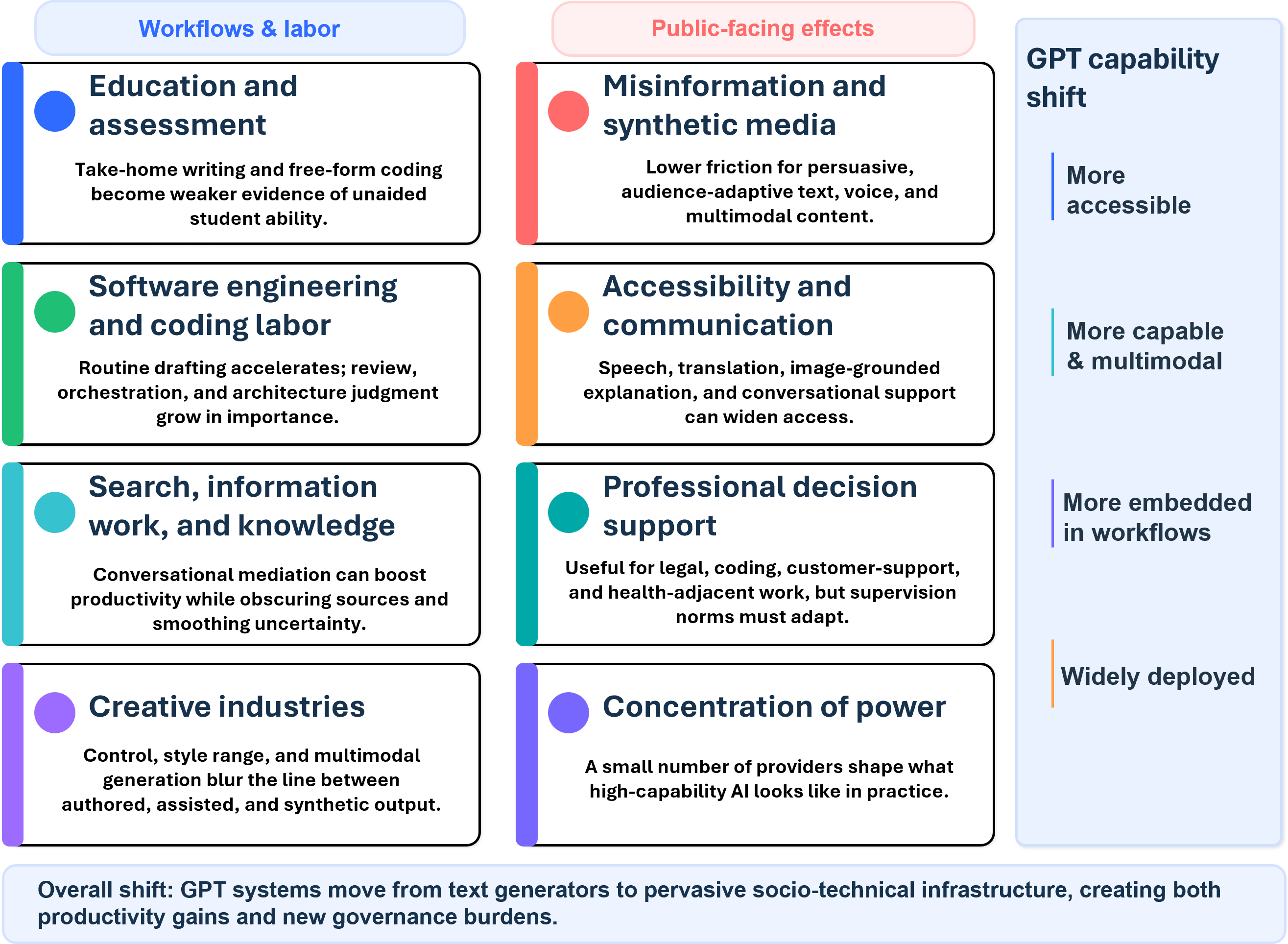}
    \caption{Societal and professional consequences: The GPT family impacted education, work, information, culture, and institutional power.}
    \label{sociatall}
\end{figure} 

\section{Societal and professional consequences}
\label{soccia}

Societal and professional consequences means what happens to how people live, work, learn, communicate, and make decisions when GPT-family systems become embedded at scale. Societal consequences refer to impacts on education, access to information, content and media production, public trust, misinformation, accessibility, and power dynamics between institutions and populations. Professional consequences refer to impacts on coding, writing, customer support, research, analysis, and decision-making. We present these consequences in Fig. \ref{sociatall} and describe them below.

\subsection{Education and assessment}

Chat-capable and ever improving GPT models have impacted education by undermining take-home writing assignments and free-form coding challenges as standalone demonstrations of student ability. Moving from GPT-3 strength in prompt engineering to GPT-3.5/GPT-4's universal accessibility allowed these models to be used by anyone for day-to-day schoolwork. The primary issue is not so much ``students are able to cheat'' but that we as educators must reorient our assessment design around process-based evidence, oral exam, authentic application, or monitored conditions.

\subsection{Software engineering and coding labor}

Later GPT generations, especially GPT-4, GPT-4.1, and GPT-5.x, materially affect software work by speeding up drafting, refactoring, code explanation, diff production, and repository navigation \cite{OpenAI2023GPT4, OpenAI2025GPT41, Fachada2025}. The very possible outcome is not simple replacement, but reallocation: more time for review, orchestration, and architectural judgment; less time for some routine synthesis.

\subsection{Search, information work, and knowledge production}

GPT systems are modifying behaviors around searching, summarization, and synthesis of information. While this can aid accessibility and productivity, it places mediation of these activities increasingly through a conversational interface that may obscure sources, collapse ambiguity, or generate smoothed-over but factually weak summaries. Harm escalates as interfaces feel more natural and are trusted more.

\subsection{Creative industries}

GPT-3 already sparked concerns about industrialization of low-cost semantic artifacts \cite{Floridi2020}. Successor models amplify this effect by increasing controllability, stylistic range, and multimodal generation and support. The result is not only economic stress to certain classes of creative labor, but increasingly obfuscated lines between authored, assisted, and synthetic outputs in workplaces.

\subsection{Misinformation and synthetic media}

As conversational fluency, multimodal capability, and persuasive realism reach new heights, risk of misinformation also evolves. GPT-4o in particular increases this attack surface with voice and multimodal interactions, while its system card claims mitigation efforts and nuanced persuasion effects \cite{OpenAI2024GPT4oSC}. The concern is no longer simply about volume of fake text. It's reducing friction for generating content that can plausibly adapt to any audience.

\subsection{Accessibility and communication}

Not all consequences are negative. Multimodal GPT systems can increase access to technology for users who need or want speech interfaces, image-grounded explanation, translation, or chat companionship. These uses have value and should not be discounted because of risk assessments.

\subsection{Professional decision support}

The family becomes increasingly capable of professional labor, such as legal, coding, customer support, and health-adjacent work \cite{OpenAI2025GPT41, OpenAI2025GPT5SC}. This enables important value but exerts stress on governance norms. The question is less whether GPTs are useful professional tools and more how those professions should recalibrate supervision, documentation, and culpability when things go wrong.

\subsection{Concentration of power}

A final consequence concerns institutional power. As highly capable GPT systems become routed, tool-using, safety-governed services, a small number of providers gain significantly high influence over what high-capability AI looks like in practice. This concentration affects research agendas, procurement decisions, platform dependency, and the politics of AI governance.

\section{Synthesis: What Actually Changed Across Generations?}
\label{synnt}

The cross-generation story can be described along five axes including nature, scope, capability, limits, and consequences.

\subsection{Nature} 

GPT-3 at its core is just a scaled few-shot text predictor. GPT-3.5 is an aligned chat assistant. GPT-4 is a more capable multimodal-input chat assistant. GPT-4o is a multimodal interaction system with real-time feedback. GPT-4.1 is presented publicly as a long-context developer oriented around tools. GPT-5.x has been publicly presented as a routed frontier workflow assistant.

\subsection{Scope} 

The scope of supported tasks expands from text-centric few-shot tasks to dialogue, multimodal interaction, large-document handling, tool use, and professional workflows. Therefore, context windows and deployment envelopes expand accordingly.

\subsection{Capability} 

Capability increases for tasks involving language, coding, longer context windows, and multimodal inputs. However, improvements are more granular and often task-specific. Generally, newer models are more capable when deployed but modeling tower is harder to directly compare due to tooling and routing taking more precedence.

\subsection{Limits} 

Hallucination, prompt fragility, evaluation instability, bias, and transparency challenges remain persistent. These problems do not go away, they just show up in different places and in different ways as systems become more complex.

\subsection{Consequences} 

AI deployed in the family goes from a testbed of few-shot scaling research questions to being infrastructural AI: built into coding tools, educational software, search, interface design, and professional workflows. The implication is that society must govern something different: not just model outputs, but deployment architectures, access mechanisms, and chains of responsibility. 

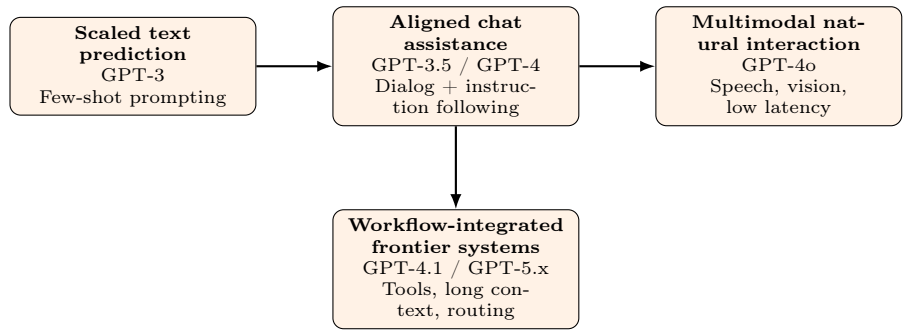
\begin{figure}[ht]
\centering
\begin{tikzpicture}[every node/.style={font=\footnotesize}]
    \node[draw, rounded corners, fill=orange!10, text width=3.0cm, align=center] (n1) {\textbf{Scaled text prediction}\\GPT-3\\Few-shot prompting};
    \node[draw, rounded corners, fill=orange!10, text width=3.0cm, align=center, right=1.0cm of n1] (n2) {\textbf{Aligned chat assistance}\\GPT-3.5 / GPT-4\\Dialog + instruction following};
    \node[draw, rounded corners, fill=orange!10, text width=3.0cm, align=center, right=1.0cm of n2] (n3) {\textbf{Multimodal natural interaction}\\GPT-4o\\Speech, vision, low latency};
    \node[draw, rounded corners, fill=orange!10, text width=3.0cm, align=center, below=1.1cm of n2] (n4) {\textbf{Workflow-integrated frontier systems}\\GPT-4.1 / GPT-5.x\\Tools, long context, routing};

    \draw[-{Latex[length=2mm]}, thick] (n1) -- (n2);
    \draw[-{Latex[length=2mm]}, thick] (n2) -- (n3);
    \draw[-{Latex[length=2mm]}, thick] (n2) -- (n4);
\end{tikzpicture}
\caption{From scaled text prediction to multimodal, tool-using, workflow-integrated frontier systems.}
\label{fig:synthesis}
\end{figure}

We present these changes across generations in Figure~\ref{fig:synthesis} considering scaled text prediction, aligned chat assistance, multimodal natural interaction, and workflow-integrated frontier systems. 

\section{Open challenges and future research directions}
\label{oppen}

Several research problems remain open and future research directions undisclosed. We present them in the form of agentic behavior, comparing proprietary models, long-context window reliability, multimodal harms, disentangling model capability, and surveying in conditions of incomplete disclosure.

\subsection{Evaluating agentic behavior} 

Later generations of GPT systems spend more time acting as components within workflows, rather than directly generating answers \cite{{X5X5}}. Evaluation must take into account tool usage, recovery from errors, quality of planning, and robustness across turns. An open challenge is that existing benchmarks emphasize single-turn correctness despite agentic systems succeeding or failing based on sequences of actions. By executing enough actions, a system can trivially seem effective due to ultimately producing correct output while planning poorly, having error-prone memory, overusing tools, or failing to recover from mistakes. Benchmarks and research task environments of the future should evaluate not just final-task success, but quality of actions, resource use, robustness to interruption, and reliability over longer interactions.

\subsection{Comparing proprietary models fairly} 

Public comparisons of proprietary models are complicated by issues such as undisclosed training data, undisclosed routing logic, renaming of models as internal code is updated, and benchmark tasks that cannot be perfectly reproduced outside of the platform. Evaluating proprietary families is itself a tool-underdeveloped area of research. Model differences may be reflective of differences in prompting, undisclosed system configuration, tool availability, or model version drift as well as differences in the underlying models. Methodological best practices for comparing proprietary models should be developed, such as standardized reporting of evaluation conditions, retesting models at different times to rule out version drift, and benchmarking protocols that attempt to factor out differences at the model-level vs platform-level. Fair comparison is thus not just a benchmarking issue but an issue of transparency and methodology.

\subsection{Measuring long-context reliability} 

Context windows near one million tokens are advertised \cite{OpenAI2026GPT41Model, OpenAI2026GPT54}, but the field still lacks stable, widely accepted measures for what models actually do with such context under realistic conditions. Reliable processing of very-long context does not imply reliable long-context reasoning, retrieval, prioritization, or constraint preservation. Models might be tricked into accepting unrealistically large contexts without actually using them, or they might be able to accept very large contexts while still hallucinating details, misweighting evidence, or failing when the task requires multi-step reasoning over segments that are far apart. Thus, going forward, research should focus less on mere context-window size claims and pay more attention to long-context use cases and realistic evaluations of these capabilities, such as retention, relevance filtering, contradiction consistency, and performance under noisy/overloaded conditions.

\subsection{Studying multimodal harms} 

Voice, image, and multimodal interaction bring with them new harms related to privacy, manipulation, identity, and grounding. These will require new longitudinal studies rather than one-off release-time audits. One challenge is that multimodal harms may arise particularly through interactions across modalities rather than due to artifacts in a single modality's outputs: e.g., voice can make systems more persuasive, images can make statements seem more credible, and paired modalities can generate more convincing illusions of grounding or social presence. Models should therefore not only be studied for accuracy, but also for how they change trust, vulnerability, persuasion, and unequal treatment over time. In particular, there is a need for longitudinal, user-centered research on these topics that studies how these harms materialize during real deployments, rather than simply under benchmark conditions.

\subsection{Disentangling model capability from product engineering} 

Later GPTs are frequently stacks of multiple models: a base model, router, safety layer, tool layer, etc. together with interface logic. Research should develop better practices to attribute which part of the stack contributed to observed successes/failures. This remains an open problem because many factors beyond changes to the base model can improve a system's real-world performance. Improvements can come from better prompting defaults or heuristics, stronger tool orchestration logic, interface guardrails or restrictions, or changes to routing policy, among other things. To better attribute progress to model improvement going forward, research should be encouraged that designs evaluation setups to more carefully isolate components. For instance, by fixing the model but training different tool logic or vice versa. Research should also be encouraged in order to probe system behavior with routing and/or scaffolding fixed. Work like this can help make claims about model progress more precise and scientifically interpretable \cite{X70}.

\subsection{Surveying in conditions of incomplete disclosure} 

AI researchers today are increasingly finding themselves studying systems for which only partial public documentation exists. In these contexts rigour demands that we analyze what we can without pretending to know what we don't. This isn't just a methodological problem; it's an epistemic one. Researchers must weigh technical reports, system cards, release notes, public evaluations, and observed behavioral while being careful to clearly state what is known, what is assumed, and what is missing. Work in this vein would benefit from better survey practices around partially disclosed systems: clearer evidential labeling, explicit accounting for uncertainty, firmer separation of documented truth versus conjecture or synthesis, and so on. In studying these systems, rigour will come not from acting like they are open but from acknowledging this fact and working around it.

\section{Conclusion}
\label{concnn}

We have examined the GPT series ranging from GPT-3 to GPT-5.x as part of a generational family. GPT-3 established large-scale few-shot prompting as a serious paradigm. GPT-3.5 transitioned that capacity towards broadly accessible conversation. GPT-4 raised the new public benchmark for general capability and safety reporting. GPT-4 Turbo expanded real-world accessibility through extended context and improved economic efficiency. GPT-4o centered real-time multimodal interactions. GPT-4.1 moved much of the center of gravity towards code, tool calling, and long-context workflows. GPT-5.x brought routing, reasoning control, and frontier workflow capabilities into the public model persona. The review also clarifies that hallucination, prompt sensitivity, variable performance, and transparency remain open challenges across the generations. To some extent, later improvements actually make evaluation more difficult: behavior comes less from static, standalone models and more from layered systems. The significant conclusion we can draw is conceptual: successive generations of GPT are not merely better versions of the same thing. They reflect a shift in the nature of AI systems’ relationship to society: from scaled text generators to increasingly multimodal, tool-using, workflow-embedded infrastructures. Therefore, future research and governance should respond accordingly.

\end{document}